\documentclass[12pt]{article}
\usepackage{enumerate}
\usepackage{natbib}
\usepackage{amsfonts,amsmath,amssymb,color,graphicx,natbib,ulem,wrapfig}
\usepackage{floatrow}
\usepackage{subfig}
\newcommand{\blind}{0}

\sffamily

\def\b0{\mathbf{0}}

\def\bA{\mathbf{A}}

\def\bI{\mathbf{I}}

\def\bI{\mathbf{I}}

\def\by{\mathbf{y}}

\def\sign{\mathrm{sign}}

\def\bx{\mathbf{x}}

\def\bX{\mathbf{X}}

\def\bX{\mbox{\bf X}}

\def\Cov{\mathrm{Cov}}

\def\Corr{\mathrm{Corr}}

\def\Var{\mathrm{Var}}

\begin{document}

\def\spacingset#1{\renewcommand{\baselinestretch}%
{#1}\small\normalsize} \spacingset{1}


\if0\blind
{
  \title{\bf Covariance-Insured Screening}
  \author{Kevin He\\
    Department of Biostatistics, School of Public Health, University of Michigan\\
        Jian Kang \\
       Department of Biostatistics, School of Public Health, University of Michigan\\
    Hyokyoung Grace Hong\\
    Department of Statistics and Probability,  Michigan State University\\
            Ji Zhu \\
    Department of Statistics, University of Michigan\\
    Yanming Li \\
    Department of Biostatistics, School of Public Health, University of Michigan\\
       Huazhen Lin \\
  School of Statistics, Southwestern University of Finance and Economics \\
  Han Xu \\
      Department of Statistics, University of Michigan\\
       and \\
    Yi Li \\
    Department of Biostatistics, School of Public Health, University of Michigan}
  \maketitle
} \fi

\if1\blind
{
  \bigskip
  \bigskip
  \bigskip
  \begin{center}
    {\LARGE\bf Modeling Time-varying Effects with Large-scale Survival
Data: An Efficient Quasi-Newton Approach}
\end{center}
  \medskip
} \fi

\bigskip
\begin{abstract}
 Modern bio-technologies have produced a vast
amount of high-throughput data with the number of predictors
far greater than the sample size. In order to identify more novel biomarkers and  understand biological mechanisms,
it is vital to detect signals weakly associated with outcomes among
 ultrahigh-dimensional predictors. However, existing screening methods, which typically ignore correlation information, are likely to miss these weak signals. By incorporating the inter-feature dependence, we propose a covariance-insured screening methodology to  identify predictors  that are jointly informative but only marginally weakly associated with outcomes. The  validity of the method is examined via extensive simulations and  real data studies for  selecting  potential genetic factors related to the onset of cancer.
\end{abstract}

\noindent%
\vfill

\newpage
\spacingset{1.45} 
\section{Introduction}
\label{sec:intro}

Rapid biological advances  have generated a vast amount
of ultrahigh-dimensional genetic data. Extracting information from these data and conducting feature selection have become a major driving force for the development of modern statistics in the last decade.

A seminal paper by \cite{Fan2008}  proposed sure independence screening (SIS) for selecting variables from ultrahigh-dimensional data. The essence  of this approach is to select variables with strong marginal correlations with the response. Much research has been inspired thereafter.  \cite{Fan2010} expanded SIS to accommodate generalized linear models,  \cite{Zhao2012} studied variable screening under the  Cox proportional hazards models, and further proposed a
score test-based screening method \citep{Zhao2014}.
Additional researches have ensured on semiparametric and nonparametric screening: semiparametric  marginal screening methods have been proposed for single-index hazard models \citep{Fan2011}, linear transformation models \citep{Zhu2011}, and general single-index models \citep{Li2012}, whereas nonparametric marginal screening methods have been proposed for linear additive models \citep{Fan2011} and  quantile regressions \citep{He2013}.

Though varied in many contexts,   these methods are based on marginal associations of individual predictors with the outcome; i.e. they assume that the true association between the individual predictors and outcomes can be inferred from their marginal associations. Although such conditions simplify theoretical derivations,  they are often violated in practice. As marginal screening methods ignore inter-feature correlations,  they tend to select irrelevant variables that are highly correlated with important variables  (false positive) and
 fail to select relevant variables that are  marginally unimportant but jointly informative  (false negative).

Because of these limitations, there has been a surge of interest in conducting multivariate screenings that  account for inter-feature dependence:   \cite{Buhlmann2010} developed a  partial correlation based algorithm (named PC-simple);  \cite{Cho2012} proposed a sequential approach (termed Tilting procedure), which measures the contribution of each variable after controlling for the other correlated variables;  \cite{Wang2016} used high-dimensional ordinary least squares projection (HOLP) that projects response to the row vectors of the design matrix, which may preserve the ranks
of regression coefficients; and \cite{Jin2014} proposed  Graphlet Screening (GS) by using the sample covariance matrix to construct a regularized graph and sequentially screening connected subgraphs.

Conceptually, multivariate screenings have been appealing. However, the computational burden increases substantially with
the number of covariates. Although simplifications have been applied to improve computational efficiency in ultrahigh-dimensional cases, they may not adequately assess the true contribution of each covariate.

For adequately assessing the association of each covariate with the response, while maintaining computational feasibility, this paper presents a new covariance-insured screening (CIS).
Leveraging the inter-feature dependence, the proposed approach is able to identify marginally unimportant but jointly informative features that are likely to be missed by conventional screening procedures. In our methodological development, we have relaxed aforementioned marginal correlation conditions that have often been assumed in the literature.  Without such restrictive assumptions, we produce the consistency results for variable selection in ultrahigh-dimensional situations.
Moreover, the proposed method is computationally efficient, and is suitable for the analysis of ultrahigh-dimensional data.

The remaining article is organized as follows:  In Section $2$, we provide some requisite preliminaries and describe our proposed method in Section 3. In Section 4 we study the theoretical properties and propose a procedure for selecting tuning parameters. Finite-sample properties are examined in Section $5$ through simulations.  We apply the proposed method to analyze a breast cancer data set in Section $6$. We conclude  with a discussion in Section $7$. All technical proofs have been deferred to the Appendix.

 \section{Related Works}
\subsection{Notation and Model}

Consider a multiple linear regression model with $n$  independent samples, $\by = \bX \boldsymbol\beta + \boldsymbol\epsilon$,
where $\by=(Y_1, \ldots, Y_n)^T$ is the response vector,  $\boldsymbol\epsilon=(\epsilon_1, \ldots, \epsilon_n)^T$ is a length-$n$ vector of independently and identically distributed random errors, $\bX$ is an $n \times p$  design matrix, and
$\boldsymbol\beta=(\beta_1, \dots, \beta_p)^T$ is the coefficient vector.
We write $\bX=[\bX_1, \ldots, \bX_n]^T=[\bx_1, \ldots, \bx_p]$, where $\bX_i$ is a $p$-dimension covariate vector for the $i$-th subject  and $\bx_j$ is the $j$-th column of the design matrix, $1 \leq i \leq n$, $1 \leq j \leq p$. Without loss of generality, we assume that each covariate $\bx_j$ is standardized to have sample mean 0 and sample standard deviation 1. For any set $\mathcal{D} \subset \{1, \dots, p\}$, we define sub-vectors, $\bX_{i, \mathcal{D}}=\{X_{i,j} : j \in \mathcal{D}\}$ and  $\bx_{\mathcal{D}}=\{\bx_j : j \in \mathcal{D}\}$.  Let $\bX_{i,-j}=\{X_{i, 1}, \ldots, X_{i, p}\} \setminus \{X_{i, j}\}$ and denote by $\boldsymbol\Sigma=\Cov(\bX_i)$.

When $p \gg n$,  $\boldsymbol\beta$ is difficult to estimate without the common sparsity condition that only a small number of variables  contribute to the response. For  improved  model interpretability and accuracy of estimation, our overarching goal is to identify the active set
\begin{align}
\mathcal{M}_0=\{j: \beta_j\neq0, ~j=1,\ldots, p\}.
\label{eq:goal1}
\end{align}

\subsection{Partial Correlation and PC-simple Algorithm}

The direct linkage between $\boldsymbol\beta$ and the partial correlations has been well established in the literature; see  \cite{Whittaker1990} and  \cite{Peng2009}, among many others.  Recently there has been much interest \citep{Buhlmann2010,Cho2012} in conducting variable screening via  partial correlations, which are defined below.

 \textbf{Definition 1}
{\it
 The partial correlation, $\rho^*(Y_i,X_{i,j}|\bX_{i,-j})$, is defined as
the correlation between the residuals resulting from the linear regression of $X_{i,j}$ on  $\bX_{i,-j}$ and $Y_i$ on $\bX_{i,-j}$}
\begin{equation} \label{eq:pc}
\rho^*(Y_i,X_{i,j}|\bX_{i,-j})=\frac{\Cov\left[Y_i-E(Y_i|\bX_{i,-j}), X_{i,j}-E(X_{i,j}|\bX_{i,-j})\right]}{\left\{\Var(Y_i-E(Y_i|\bX_{i,-j}))\Var(X_{i,j}-E(X_{i,j}|\bX_{i,-j}))\right\}^{1/2}}.
\end{equation}

When $p$ is large,  estimating  partial correlations is computationally cumbersome.  \cite{Buhlmann2010} have proposed a PC-simple algorithm to compute  lower-order partial correlations $\rho^*(Y_i, X_{i,j}|\bX_{i, \mathcal{C}})$ sequentially for some $\mathcal{C} \subseteq \{1, \ldots, p\} \setminus \{j\}$ with the cardinality $||\mathcal{C}||_0=0, \dots, m$, where $m$ is a pre-specified integer. When $m=0$ or $\mathcal{C}$ is empty, the PC-simple algorithm is a special case of  the SIS procedure.

The PC-simple algorithm avoids the computation of high-order partial correlations and provides a new approach for variable screening. The validity of this algorithm hinges upon the condition that $
  \rho^*(Y_i, X_{i,j}|\bX_{i,\mathcal{C}})=0$ implies $\rho^*(Y_i, X_{i,j}|\bX_{i,-j})=0$.
To examine this condition, \cite{Cho2012} considered a sample version of (\ref{eq:pc})
   \begin{align*}
\hat\rho^*(Y_i,X_{i,j}|\bX_{i,\mathcal{C}})=\frac{\bx_j^T(\bI_n-\Pi_{\mathcal{C}})\by}{\sqrt{\bx_j^T(\bI_n-\Pi_{\mathcal{C}})\bx_j}\sqrt{\by^T(\bI_n-\Pi_{\mathcal{C}})\by}},
\end{align*}
 where $\bI_n$ is the identity matrix and $\Pi_{\mathcal{C}} =\bx_{\mathcal{C}}(\bx_{\mathcal{C}}^T\bx_{\mathcal{C}})^{-1}\bx_{\mathcal{C}}^T $ is the projection matrix onto the space  spanned by $\bx_{\mathcal{C}}$.  The numerator of $\hat\rho^*(Y_i,X_{i,j}|\bX_{i,\mathcal{C}})$
can be decomposed as
\begin{align}
\bx_j^T(\bI_n-\Pi_{\mathcal{C}})\by=\beta_j\bx_j^T(\bI_n-\Pi_{\mathcal{C}})\bx_j+\sum_{k \in \mathcal{M}_0 \setminus (\mathcal{C}\cup\{j\})} \beta_k\bx_j^T(\bI_n-\Pi_{\mathcal{C}})\bx_k+\bx_j^T(\bI_n-\Pi_{\mathcal{C}})\boldsymbol\epsilon.
\label{eq:equation4}
\end{align}
Equation \eqref{eq:equation4} indicates that  only when  the last two terms on the right hand side of
(\ref{eq:equation4}) are negligible compared to the first one,  the PC-algorithm is valid  and  $\hat\rho^*(Y_i,X_{i,j}|\bX_{i, \mathcal{C}})$
 can be used to identify $\mathcal{M}_0$ in lieu of  $\hat\rho^*(Y_i,X_{i,j}|\bX_{i, -j})$.  In practice, however, there is no guarantee this condition would hold for an arbitrary set $ \mathcal{C}$.

\subsection{Tilting Procedure}

As a remedy, the Tilting procedure \citep{Cho2012} identified a data-driven conditioning set  $\mathcal{C}$. Specifically, for each variable under consideration, the corresponding $ \mathcal{C}$ contains all variables that are highly correlated with it. While successful, this procedure also has unsolved issues. For instance, this way of selecting $ \mathcal{C}$ may not adequately assess the true contribution of each covariate, As a result, important predictors that have weak marginal effects but strong joint effects can be missed (a simple example is provided in the Supplementary Materials). Moreover, the computational cost  grows drastically with the number of predictors. New methodologies are needed for  adequately assessing the true contribution of each covariate, while maintaining computational feasibility. These concerns motivate the proposed method.

\section{Proposed Method}

As discussed previously, a well-constructed $\mathcal{C}$ is critical. We propose compartmentalizing covariates into blocks so that variables from distinct blocks are less correlated. This solution may bypass the difficulty encountered in existing multivariate screening procedures and   render improved computational feasibility, better screening efficiency and weaker theoretical conditions. Our proposal is detailed below.

\subsection{Preamble}

First,  in order to identify the active set $\mathcal{M}_0$,
we  connect  $\beta_j$ to   the semi-partial correlation \citep{Kim2015}, a modified version of  partial correlation that is defined below.

 \textbf{Definition 2}
{\it
 The semi-partial correlation, $\rho(Y_i,X_{i,j}|\bX_{i,-j})$, is defined as
the correlation between $Y_i$ and the residuals resulting from the linear regression of $X_{i,j}$ on  $\bX_{i,-j}$, i.e.}
\begin{equation} \label{eq:spc}
\rho(Y_i,X_{i,j}|\bX_{i,-j})=\frac{\Cov\left[Y_i, X_{i,j}-E(X_{i,j}|\bX_{i,-j})\right]}{\left\{\Var(Y_i)\Var(X_{i,j}-E(X_{i,j}|\bX_{i,-j}))\right\}^{1/2}}.
\end{equation}

Indeed, the following lemma reveals  that $\rho(Y_i,X_{i,j}|\bX_{i,-j})$ infers the effect of $X_{i,j}$ on $Y_i$ conditional on  $\bX_{i,-j}$ and hence  identifying  \eqref{eq:goal1} is equivalent to finding
 \begin{align}
\mathcal{M}_0=\{j: \rho(Y_i,X_{i,j}|\bX_{i,-j}) \neq 0, ~j=1,\ldots, p\}.
\label{eq:goal2}
\end{align}

 \textbf{Lemma 1}
{\it
Suppose that $\boldsymbol\Sigma$ is  positive definite. Then
\[
\beta_j =0 ~\mbox{if~and~only~if}~ \rho(Y_i,X_{i,j}|\bX_{i,-j})=0.
\]}

The intuitions of the proposed covariance-insured screening method are further provided by the following lemma.

 \textbf{Lemma 2}
{\it Suppose that the predictors can be partitioned into independent blocks, $\mathcal{S}_1, \ldots, \mathcal{S}_G$.  For any $j=1, \ldots, p$ and some $g$ such that $j \in \mathcal{S}_g$,  \[
 \rho(Y_i,X_{i,j}|\bX_{i,-j})=\rho(Y_i, X_{i, j} |  \bX_{i, \mathcal{S}_g \setminus \{j\}}).
\]}

We first note that the equality in Lemma 2 does not hold for partial correlations, which motivates the use of semi-partial correlations instead.
Second,  Lemma 2 provides the intuition behind the proposed method. However, the independent block assumption is not required for the proposed method, which is valid for more general settings by thresholding the sample covariance matrix \citep{Bickel2008} and compartmentalizing covariates  into less correlated blocks. Constructing less correlated blocks is well understood in  genetics literature and  is often of interest per se \citep{Berisa2016}. For example, in a cutaneous melanoma study \citep{He2016}, 2,339 single-nucleotide polymorphisms (SNPs)  could be  grouped into 15 blocks; see Figure 1.

\subsection{Thresholding Sample Covariance Matrix}

To formalize the idea of thresholding, consider $\widehat{\boldsymbol\Sigma}$ the sample estimate of $\boldsymbol\Sigma$. For a threshold $\delta > 0$,  let $\widehat{\boldsymbol\Sigma}^{\delta}$ be the regularization of $\widehat{\boldsymbol\Sigma}$ such that
   \begin{align*}
   \widehat{\boldsymbol\Sigma}^{\delta}_{jk} = \widehat{\boldsymbol\Sigma}_{jk} 1\{|\widehat{\boldsymbol\Sigma}_{jk}| \geq \delta\}.
    \end{align*}
 We then partition the vector $\boldsymbol\beta$ into blocks, $\hat{\mathcal{S}}_1, \ldots, \hat{\mathcal{S}}_G$,  in a way such that  all off-diagonal blocks of $\widehat{\boldsymbol\Sigma}^{\delta}$ are zero;
e.g.
   \begin{align*}
  |\widehat{\boldsymbol\Sigma}^{\delta}_{jk}| = 0 ~ for ~all ~ j \in \hat{\mathcal{S}}_g, ~k \in \hat{\mathcal{S}}_{g'}, ~g \neq g'.
   \end{align*}
Here $\hat{\mathcal{S}}_1, \ldots, \hat{\mathcal{S}}_G$ forms a partition of the $p$ predictors:
   \begin{align*}
\hat{\mathcal{S}}_g \cap \hat{\mathcal{S}}_{g'} = \varnothing ~for ~g \neq g', ~and~ \hat{\mathcal{S}}_1 \cup \hat{\mathcal{S}}_2 \ldots \cup \hat{\mathcal{S}}_G = \{1, \ldots, p\}.
     \end{align*}
Further details for the partition and recommendations for the choice of $\delta$ are provided in the Supplementary Materials.

\subsection{Covariance-Insured Screening}

With the  block diagonal $\widehat{\boldsymbol\Sigma}^{\delta}$, the disconnected blocks are approximately orthogonal, which motivates
us to fit block-wise procedures to compute the semi-partial correlation within each identified block.
The proposed approach can be summarized as follows:

  \begin{enumerate}
 \item[Step 1:] Identify the disconnected blocks  by thresholding the sample covariance matrix.
 \item[Step 2:] Compute the block-wise sample semi-partial correlations $\hat\rho(Y_i,X_{i,j}|\bX_{i,\hat {\mathcal{S}}_g \setminus \{j\}})$. For each $j \in \hat {\mathcal{S}}_g$, $1\leq g \leq G$, denote
 $\Pi_{\hat{\mathcal{S}}_g\setminus\{j\}}$ as the projection matrix onto the space  spanned by $\bx_{\hat {\mathcal{S}}_g \setminus \{j\}}$. That is,
   \begin{align*}
\Pi_{\hat{\mathcal{S}}_g\setminus\{j\}}=\bx_{\hat {\mathcal{S}}_g \setminus \{j\}}(\bx_{\hat {\mathcal{S}}_g \setminus \{j\}}^T\bx_{\hat {\mathcal{S}}_g \setminus \{j\}})^{-1}\bx_{\hat {\mathcal{S}}_g \setminus \{j\}}^T.
\end{align*}
 Then the block-wise sample semi-partial correlation  can be calculated as
   \begin{align*}
\hat\rho(Y_i,X_{i,j}|\bX_{i, \hat {\mathcal{S}}_g \setminus \{j\}})=\frac{\bx_j^T(\bI_n-\Pi_{\hat{\mathcal{S}}_g\setminus\{j\}})\by}{\sqrt{\bx_j^T(\bI_n-\Pi_{\hat{\mathcal{S}}_g\setminus\{j\}})\bx_j}\sqrt{\by^T\by}}.
\end{align*}
 \item[Step 3:] Compute
\begin{align*}
  \widehat{\mathcal{M}}_{CIS}=\left\{j \in \hat{\mathcal{S}}_g,~ 1\leq g \leq G:~\left|\hat{\rho}(Y_i,X_{i,j}|\bX_{i, \hat {\mathcal{S}}_g \setminus \{j\}})\right| > \nu\right\},
\end{align*}
where $\nu$ is a pre-defined threshold.
\end{enumerate}

\section{Asymptotic Property for CIS}

\subsection{Conditions and Assumptions}

To make our results general,  we allow the dimension of covariates  and the active set to grow as a function of sample size (i.e. $p=p_n$), and  $\mathcal{M}_0=\mathcal{M}_{0,n}$. Under some commonly assumed conditions below,
we show that the CIS procedure identifies the true active set $\mathcal{M}_{0,n}$ with a probability tending to 1.

 \begin{enumerate}
 \item[(A1)] $||\mathcal{M}_{0,n}||_0=O(n^a)$ for some $a \in [0, \frac{1}{2})$, where $||\cdot||_0$ denotes the cardinality.
     \item[(A2)] The dimension of the covariates is  $p_n=\exp(n^c)$ for some $ c \in [0, 1-2b)$, where  $b \in (a, \frac{1}{2})$.
 \item[(A3)]  For a threshold $\delta_n=O(\sqrt{\log(p_n)/n})$,  $q_n=\max_{1 \leq g \leq G_n} ||\hat {\mathcal{S}}_g||_0$, the maximal number of variables in the disconnected blocks satisfies $q_n\leq C_1n^{d}$ for some constant $C_1>0$ and $d \in [0, b-a)$, where $G_n$ is the number of blocks (depending on $n$).
     \item[(A4)] Let $\lambda_{max}(\bA)$ and $\lambda_{min}(\bA)$ represent the largest and smallest eigenvalues of an arbitrary positive definite matrix $\bA$. There exist two positive constants $\tau_{min}$ and $\tau_{max}$ such that
          \begin{align*}
0<\tau_{min} \leq \min_{\mathcal{D}}\{\lambda_{min}(\frac{1}{n}\bx_{\mathcal{D}}^T\bx_{\mathcal{D}})\} \leq \max_{\mathcal{D}}\{\lambda_{max}(\frac{1}{n}\bx_{\mathcal{D}}^T\bx_{\mathcal{D}})\} \leq \tau_{max}<\infty
 \end{align*}
for any $\mathcal{D} \subset \{1, \dots, p\}$ with cardinality $||\mathcal{D}||_0 \leq n^{\max\{a,d\}}$.
   \item[(A5)]  Assume non-zero coefficients $\beta_j$ satisfying
  $\max_{j \in  \mathcal{M}_{0,n}}|\beta_j|<M$ for some $M \in (0, \infty)$ and $n^\kappa \min_{j \in  \mathcal{M}_{0,n}}|\beta_j|\rightarrow \infty$ for $\kappa \in [0, b-a-d)$.
  \item[(A6)]    Assume that $\max_{1\leq j \leq p} ||\bx_j||_{\infty}\leq K$ for a constant $K>0$, where $||\bx_j||_{\infty}=\max_{1\leq i \leq n} |X_{i, j}|$.
  \item[(A7)]    The random errors follow a sub-exponential distribution; i.e. $\epsilon_1, \ldots, \epsilon_n$ are independent random variables with mean 0 and satisfy
  \begin{align*}
\frac{1}{n}\sum_{i=1}^n E|\epsilon_i|^m \leq \frac{m!}{2} K_{\epsilon}^{m-2}, ~m=2,3, \ldots,
\end{align*}
where $K_{\epsilon}$ is a constant depending on the distribution.
\end{enumerate}
\vspace{-\topsep}
Condition (A1) allows the number of non-zero coefficients $||\mathcal{M}_{0,n}||_0$ to grow with the sample size $n$. Condition (A2) allows for an exponential growth of dimension as a function of sample size, i.e. ultrahigh-dimensionality.   The bound of $q_n$ in Condition (A3) guarantees the existence of the projection matrix $\Pi_{\hat{\mathcal{S}}_g\setminus\{j\}}$ and hence the corresponding sample semi-partial correlations. Condition (A4) rules out the strong collinearity between variables, which was also assumed in \cite{Candes2007, Zhang2008, Wang2009, Cho2012}. As shown in \cite{Wang2009}, there is a connection between Condition (A4) and the condition requiring strict positive definiteness of the population covariance matrix, which is commonly  assumed in the variable selection literature \citep{Fan2001, Zou2006, Buhlmann2010}. For instance, when both $\bX$ and $\boldsymbol\epsilon$ follow the normal distribution, the former is implied by the latter.
 Condition  (A5)  controls the magnitude of the non-zero coefficients, which  was also assumed in \cite{Fan2008} and \cite{Wang2016}. Condition (A6) is usually satisfied in practice, and it is helpful for the proof of probability inequalities.
The sub-exponential distribution in Condition (A7) is general and includes many commonly assumed distributions.

\subsection{Main Theorem}

We establish the important properties of CIS by presenting the following theorem.

\textbf{Theorem 1 (screening consistency)}
{\it Assume that (A1)-(A7) hold. Denote by $\widehat{\mathcal{M}}_{CIS}(\nu_{n}, \delta_n)$ the set of selected variables from the CIS procedure in Section 3.3 with the tuning parameters $\nu_{n}=n^{-\kappa}$ and $\delta_n=O(\sqrt{\log(p_n)/n}~)$. Then the following two statements are true.
 \begin{align*}
&{} P\left(\min_{j \in \hat {\mathcal{S}}_g \cap\mathcal{M}_{0,n}, ~g=1, \ldots, G_n} \nu_n^{-1}|\hat\rho(Y_i,X_{i,j}|\bX_{i, \hat {\mathcal{S}}_g \setminus \{j\}})| \geq O(n^{\kappa}\min_{j \in  \mathcal{M}_0}|\beta_j|)\right) \geq 1-O\left(\exp\left(-n^{1-2b}\right)\right),
\end{align*}
 \begin{align*}
&{} P\left(\max_{j \in \hat {\mathcal{S}}_g \setminus\mathcal{M}_{0,n}, g=1, \dots, G_n}\nu_n^{-1}|\hat\rho(Y_i,X_{i,j}|\bX_{i, \hat {\mathcal{S}}_g \setminus \{j\}})|\leq O(n^{-(b-a-d/2-\kappa)})\right)
 \geq 1-O\left(\exp\left(-n^{1-2b}\right)\right).
\end{align*}
These results imply the screening consistency property that
  \begin{align*}
P(\widehat{\mathcal{M}}_{CIS}(\nu_{n}, \delta_n)=\mathcal{M}_{0,n}) \geq 1-O\left(\exp\left(-n^{1-2b}\right)\right).
\end{align*}
}
\vspace{-10mm}

\subsection{Selection of Tuning Parameters}

Even though  theoretical thresholds have been derived  in various variable screening procedures, it remains a challenge to implement them. Moreover, strong correlations among predictors may deteriorate the performance of screening procedures in finite samples.
To address these challenges, iterative SIS (ISIS) \citep{Fan2008} was proposed as a  remedy for marginal screening procedures.
 Along the same lines,
we design an iterative CIS algorithm (termed ICIS) and
further build a thresholding procedure to control  false discoveries.

  \begin{enumerate}
 \item[Step 1:] Resample the original data with replacement multiple (say $B$) times.
  \item[Step 2:] For each resampled data, first identify the variables by the proposed CIS procedure, followed by applying adaptive Lasso for  variable selection and computing  the associated residuals in the regression.
    \item[Step 3:] Treating those residuals as new responses, we apply CIS  to the remaining variables.
    \item[Step 4:] We repeat the procedure until a pre-specified number of iterations is achieved or the selected variables do not change.
\item[Step 5:] Denote the selected variable index set from the $r$-th resampled data as $\widehat{\mathcal{M}}^{(r)}$ for $r=1,\ldots, B$.  Let $\widehat \Psi_j$ be the empirical  probability that the $j$-th variable is selected; i.e.
    \begin{eqnarray}
\widehat \Psi_j=\frac{1}{B}\sum_{r=1}^B I(j \in \widehat{\mathcal{M}}^{(r)}).
\nonumber
\end{eqnarray}
For a threshold $\psi \in (0,1)$, the procedure selects  variables with
    \begin{align}
\widehat{\mathcal{M}}_{ICIS}=\{j: \widehat\Psi_j\geq \psi, ~j=1,\ldots,p\}.
\label{eq:CIS}
\end{align}
\end{enumerate}

To  determine data-driven thresholds for the selection frequency $\psi$, we further adopt a random permutation-based approach \citep{He2016} to control the empirical Bayes false discovery rate \citep{Efron2012}.
For a pre-specified value $q \in (0,1)$,  $\psi$ will be chosen to ensure that at most  $q$ proportion of the selected variables would be false positives. Further technical details are provided in the Supplementary Materials.

\section{Simulation Study}

\subsection{Performance of the CIS}

We assess the performance of  the proposed CIS method by comparing it with SIS, non-iterative versions of HOLP and the Tilting procedure under various simulation configurations. For each  configuration a total of  100 independent data are  generated.

 \begin{enumerate}
  \item[(Model A)] Data are generated with $n=1,000$ and $p=10,000$, from a
multivariate normal distribution with a block-diagonal  covariance structure ($m=100$ independent blocks, each with  100 predictors). Within each block the variables  follow a AR1 model with the auto-correlation varying from 0.5 to 0.9. The variables with non-zero effects are
\begin{align*}
X_1,~X_2,~X_{m+1},~X_{m+2},~X_{2m+1},~X_{3m+1},~X_{4m+1},~X_{5m+1},~X_{6m+1},~X_{7m+1}
\end{align*}
  with the corresponding coefficients  $1,-1,1,-1,-1,1,-1,1,-1,1$.

\item[(Model B)]  This model is similar to Model A, but the  variables with non-zero effects are
   \begin{align*}
  X_1,~X_{m+1},~X_{2m+1},~X_{3m+1},~X_{4m+1},~X_{5m+1},~X_{6m+1},~X_{7m+1}~X_{8m+1},~X_{9m+1}
   \end{align*}
    with the corresponding coefficients  $1,1,-1,1,-1,1,-1,1, -1, 1$.

    \item[(Model C)]  This model is similar to Model A, but the covariance matrix is not block-diagonal (e.g. all variables belong to the same block).
The variables with non-zero effects are
\begin{align*}
X_{j_1},~X_{j_1+1},~X_{j_2},~X_{j_2+1},~X_{j_3},~X_{j_4},~X_{j_5},~X_{j_6},~X_{j_7},~X_{j_8}
\end{align*}
  with the corresponding coefficients  $1,-1,1,-1,-1,1,-1,1,-1,1$, where the indices $j_1, \ldots, j_8$ are  randomly drawn from $\{1, \ldots, p\}$.
 \end{enumerate}

Table 1 compares   the minimum number of selected variables to include the true  model (Model Size). For Model A and B,  the threshold $\delta_n=5\sqrt{\log(p)/n}$ is used to determine blocks in the CIS procedure. Since Model C  does not have a block-diagonal covariance structure, we  partition variables into 10 blocks in which variables with high absolute correlations reside.  When the within-block correlations are low, all methods perform well and the average model size is close to 10, the true model size. When the correlations  are high (e.g. greater than 0.6),  CIS
outperforms  SIS and Tilting in the presence of signal cancelations (Models A and C). The poor performance of  SIS and Tilting can be explained in part because the strong marginal correlation condition is not satisfied. Interestingly, HOLP is competitive and  performs well for Model B and C, but does not work well for Model A. This might be caused by the violation of the diagonal dominance of $\bX^T(\bX\bX^T)^{-1}\bX$ required by HOLP.

\subsection{Performance of the iterative CIS (ICIS)}

We compare ICIS with Lasso, adaptive Lasso, ISIS, iterative HOLP and Tilting.

 \begin{enumerate}
  \item[(Model D)] This model is similar to Model A. Within each block, the variables  follow a AR1 model with parameter 0.9. The effect size is chosen as  $\beta=(0.5, 0.75, 1)$ to generate  a wide range of  signal strength.
 \end{enumerate}

 Data are generated with $p=1,000$ or $10,000$. No
results are reported for Tilting  with $p=10,000$ due to its intensive computation.  As indicated in Table 2, iterative CIS outperforms most  methods, yielding the smallest false negative (FN) and  false positive (FP) combined.

We next compare the proposed method with iterative Graphlet Screening (termed GS). We consider Experiment 2b reported in \cite{Jin2014}, which is described as follows:

 \begin{enumerate}
  \item[(Model E)] Data are generated with $p=5,000$ and $n=p^{\kappa}$ with $\kappa=0.975$. We consider the following Asymptotic Rare and Weak (ARW) model \citep{Jin2014}. The signal vector $\boldsymbol\beta$ is modeled by $\boldsymbol\beta=\textbf{b}\circ \boldsymbol\mu$, where $\circ$ denotes the Hadamard product. The vector of $\boldsymbol\mu$ consists of    $z_j|\mu_j|$, $j=1, \ldots, p$, where $z_j=\pm 1$ with equal probability and  $|\mu_j| \sim 0.8\nu_{\tau_p}+0.2h$, where $\nu_{\tau_p}$ is the point mass at $\tau_p$ with $\tau_p=\sqrt{6\log(p)}$. The $h(x)$ is the density of $\tau_p(1+V/6)$, $V\sim\chi_1^2$. We choose the correlation matrix to be a diagonal block matrix where each block is a 4 by 4 matrix satisfying $\Corr(X_{i,j}, X_{i,j'})=I(j=j')+0.4I(|j-j'|=1)\times \sign(6-j-j')+0.05I(|j-j'|>2)\times\sign(5.5-j-j')$, $1\leq j,j'\leq 4$. The vector $\textbf{b}$ consists of $b_j$, $j=1, \ldots, p$, where $b_j=0$ or $1$. Let $k$ be the number of variables with $b_j \ne 0$ within each block. With  $\pi=0.2$ and $\vartheta=0.35$, we randomly choose $(1-4p^{-\vartheta})$ fraction of the blocks for $k=0$ (e.g. $b_j=0$ for all $j$ belongs to these block), $4(1-\pi)p^{-\vartheta}$ fraction of the blocks for $k=1$, and $4\pi p^{-\vartheta}$ fraction of the block for $k \in \{2, 3, 4\}$.
 \end{enumerate}

Table 3 compares  Lasso, adaptive Lasso, ISIS, ICIS and GS.  No results are reported for HOLP (intensive computation for large $n$) or Tilting (intensive computation for large $p$). Web Figure A1 in the Supplementary Materials compares ICIS and GS with various choices of tuning parameters. The results suggest that the perturbation of tuning parameters has relatively small effects on the proposed ICIS, which outperforms GS.

\section{Real Data Study}

\subsection{Multiple Myeloma Data}

Multiple myeloma (MM) represents more than 10 percent of all
hematologic cancers in the U.S. \citep{Kyle}, resulting in more
than 10,000 deaths each year. Developments in gene-expression
profiling and sequencing of MM patients have offered effective
ways of understanding the cancer genome \citep{Chapman}.  Despite
this promising outlook, analytic methods remain insufficient for
achieving truly personalized medicine. The standard procedure is to evaluate one gene at a time, which results in low statistical power to identify the disease-associated genes \citep{Sun}.
Thus, more accurate models that leverage the large amounts of
genomic data now available are in great demand.

Our goal is to identify genes that are relevant to  the Beta-2-microglobulin (Beta-2-M), which is a continuous prognostic factor for multiple myeloma. We use gene expression and Beta-2-M from 340
 multiple myeloma patients who were recruited into clinical trial UARK 98-026,
which studied total therapy II (TT2). These data are described in  \cite{Shaughnessy}, and can be obtained through the MicroArray Quality Control Consortium II study \citep{MAQC}, available on GEO (GSE24080). Gene expression profiling was performed using Affymetrix U133Plus2.0 microarrays. Following the strategy in \cite{Zhao2014},
 we averaged the expression levels of probesets corresponding to the same gene, resulting in 20,502 covariates.

\subsection{Analysis Methods}

 For discovering genetic variants relevant to the risk of cancer, an often ignored fact  is that the genetic variants possess block correlation structures. In our motivating MM study, the estimated correlation matrix of gene expressions is  nearly block diagonal under a suitable permutation of the variables. The predictors are strongly correlated within blocks
and are less correlated between blocks (a sample correlation plot is shown in Figure 2a). Hence, many elements of the covariance matrix are small.
 A major challenge, arising from such a correlation structure, is that some genes can be jointly relevant but not marginally relevant to the disease outcome. These genes are often termed as "hidden" since the random noise manifested in the data often obscures their impact.
In such a
difficult setting, popular methods such as marginal screening and multivariate screening
are simply overwhelmed. The marginal screening is overwhelmed as it largely neglects
correlations across predictors. Exhaustive multivariate screening is overwhelmed as it is
neither computationally feasible nor efficient.

 To select the informative genes, the proposed ICIS is implemented on the MM data set with 340 subjects. The thresholding parameter for $\delta_n$ is fixed at $5\sqrt{\log(p)/n}$ such that the maximal number of variables in the disconnected blocks satisfies $q_n\leq n$. The importance of predictors is evaluated
 by the selection frequencies among the $50$ resampled data.
The estimated false discovery rate is calculated to determine a data-driven threshold $\psi$ (defined in Section 4.3) for the selection frequency such that at most $q$ proportion of the selected variables would be false positives. We compare the ICIS with Lasso, adaptive Lasso, ISIS and iterative HOLP.

\subsection{Results}

Using our method, a total of $24$ genes pass the threshold  for $q=0.1$.
In comparison, the Lasso, adaptive Lasso, ISIS and HOLP procedures select $74, 0, 0$ and $54$ genes, respectively.
All these results are consistent with
those from the Simulation section. The Lasso tends to select many
irrelevant variables, while the adaptive Lasso and ISIS suffer from a reduced
power to identify informative predictors. The proposed method
selects substantially fewer variables than the HOLP and provides a
control for false discoveries. Some of the genes selected by the proposed method confirm those identified by Lasso and HOLP.
One of the top common genes, MMSET (multiple myeloma SET domain containing
protein), is known as the key molecular target in MM \citep{Mirabella} and has been involved in the chromosomal translocation
in MM. Another selected gene, FAM72A (Family With Sequence
Similarity 72 Member A), has been reported to be associated with
poor prognosis in multiple myeloma \citep{Noll}. Moreover,
expression level of gene ATF6 (Activating Transcription Factor 6)
has been reported to predict the response of multiple myeloma
to the proteasome inhibitor Bortezomib \citep{Nikesitch}.

In addition, besides confirming  genes already selected by the competing methods, we also find
novel signals. For example, among the genes in our finding but not in other methods, Phospholipase C epsilon 1 (PLCE1), EH-domain containing 2 (EHD2), long intergenic non-protein coding RNA 665 (LINC00665) and ZNF295 Antisense RNA 1 (ZNF295-AS1)
 are correlated with each other (see Figure 2b) and have reversed covariate effects (-0.51, 1.02, -0.23 and -0.43). These results suggest the existence of signal cancelations. The failure of identifying such genes by other screening methods  may be explained in part
because the strong marginal correlation condition is not satisfied. In fact,
these genes are likely to be associated with the prognostic of the MM, as reported by previous literature. For instance,
PLCE1, located on chromosome 10q23, encodes a phospholipase that has been reported to be associated with intracellular signaling through the regulation of a variety of proteins such as the protein kinase C (PKC) isozymes and the proto-oncogene ras \citep{Rhee2001,Bunney2009}. On the other hand,
EDH2 is a plasma membrane-associated member of the EHD family, which regulates internalization and is related to actin cytoskeleton. Abnormal expression of EHD2 has been linked  to metastasis of carcinoma \citep{Li2013}. In addition,  \citet{Zhang2016} suggested linc00665  might play a role as "sponge" to indirectly de-repress a series of mRNAs in nasopharyngeal nonkeratinizing carcinoma. It appears that
the proposed approach leverages the dependence among covariates and is able to identify jointly-informative variables that only have marginally weak associations with outcomes.

\section{Discussion}
\label{s:discuss}
We have developed a covariance-insured  screening method for ultrahigh-dimensional variables.
The innovation lies in that, as opposed to conventional variable screening methods, the proposed approach
leverages the dependence structure among covariates and is able to identify jointly informative variables that only have weak marginal
 associations with outcomes. Moreover, the proposed method is computationally efficient, and thus suitable for the analysis of ultrahigh-dimensional data. Finally, the proposed CIS procedure can be extended to accommodate non-linear models. We will report these extensions elsewhere.

\section*{Appendix}

\subsection*{Proof of Theorem 1}

The block-wise sample semi-partial correlation can be calculated as
   \begin{align*}
\hat\rho(Y_i,X_{i,j}|\bX_{i, \hat {\mathcal{S}}_g \setminus \{j\}})=\frac{\frac{1}{n}\bx_j^T(\bI_n-\Pi_{\hat {\mathcal{S}}_g \setminus \{j\}})\by}{\frac{1}{n}\sqrt{\bx_j^T(\bI_n-\Pi_{\hat{\mathcal{S}}_g \setminus \{j\}})\bx_j}\sqrt{\by^T\by}},
\end{align*}
where $j \in \hat {\mathcal{S}_g}$ for some $g$, a factor $1/n$ is applied to both numerator and denominator to facilitate  asymptotic derivations, and $\Pi_{\hat{\mathcal{S}}_g\setminus\{j\}}$ is the projection matrix onto the space  spanned by $\bx_{\hat {\mathcal{S}}_g \setminus \{j\}}$
   \begin{align*}
\Pi_{\hat{\mathcal{S}}_g\setminus\{j\}}=\bx_{\hat {\mathcal{S}}_g \setminus \{j\}}(\bx_{\hat {\mathcal{S}}_g \setminus \{j\}}^T\bx_{\hat {\mathcal{S}}_g \setminus \{j\}})^{-1}\bx_{\hat {\mathcal{S}}_g \setminus \{j\}}^T.
\end{align*}
Because
$\beta_k\bx_j^T(\bI_n-\Pi_{\hat{\mathcal{S}}_g \setminus \{j\}})\bx_k=0$ for $k \in \mathcal{M}_{0,n} \cap (\hat{\mathcal{S}}_g \setminus \{j\})$,
the numerator can be decomposed as
   \begin{equation}
 \frac{1}{n}\bx_j^T(\bI_n-\Pi_{\hat{\mathcal{S}}_g \setminus \{j\}})\by
 =\frac{1}{n}\beta_j\bx_j^T(\bI_n-\Pi_{\hat{\mathcal{S}}_g \setminus \{j\}})\bx_j+\frac{1}{n}\sum_{k \in \mathcal{M}_{0,n} \setminus \hat{\mathcal{S}}_g} \beta_k\bx_j^T(\bI_n-\Pi_{\hat{\mathcal{S}}_g \setminus \{j\}})\bx_k  +\frac{1}{n}\bx_j^T(\bI_n-\Pi_{\hat{\mathcal{S}}_g \setminus \{j\}})\boldsymbol\epsilon. \label{eq:5}
\end{equation}

To show the first and the second statements in Theorem 1, we consider the following two scenarios respectively:
(1) $j \in \mathcal{M}_{0,n}$ and (2)
 $j \in \{1, \dots, p_n\} \setminus \mathcal{M}_{0,n}$.

{\bf Step 1: $j \in \mathcal{M}_{0,n}$}

{\bf Step 1.1} We first aim to show that for $j \in \mathcal{M}_{0,n}$ the absolute value of the first term  on the right hand side of \eqref{eq:5} can be bounded from below and the last two terms on the right hand side of \eqref{eq:5} are negligible compared to the first term.

Specifically, for  the first term we can show that for some $g$ such that $j \in \hat {\mathcal{S}_g}$,
  \begin{align*}
\min_{j \in \hat {\mathcal{S}}_g \cap\mathcal{M}_{0,n}, ~g=1, \ldots, G_n}\Big|\frac{1}{n}\beta_j\bx_j^T(\bI_n-\Pi_{\hat{\mathcal{S}}_g \setminus \{j\}})\bx_j\Big|\geq \alpha\min_{j \in \mathcal{M}_{0, n}}|\beta_j|,
\end{align*}
where $\alpha>0$ is a given constant.

By the  property of the determinant of a partitioned matrix, when $\textbf{A}$ is   non-singular,
\[ \det\left ( \begin{array}{cc}
\textbf{A} & \textbf{B}  \\
\textbf{C} & \textbf{D}  \end{array} \right) =\det(\textbf{A})\det(\textbf{D}-\textbf{C}\textbf{A}^{-1}\textbf{B}).\]
Then we have
\begin{align*}
\det\left(\frac{1}{n}\bx_{\hat {\mathcal{S}}_g}^T\bx_{\hat {\mathcal{S}}_g}\right)=&{}
\det\left(\frac{1}{n}\bx_{\hat {\mathcal{S}}_g\setminus \{j\}}^T\bx_{\hat {\mathcal{S}}_g\setminus \{j\}}\right) \det\left(\frac{1}{n}\bx_j^T\bx_j-\frac{1}{n}\bx_j^T\Pi_{\hat {\mathcal{S}}_g \setminus \{j\}}\bx_j\right)\\
=&{}
\det\left(\frac{1}{n}\bx_{\hat {\mathcal{S}}_g\setminus \{j\}}^T\bx_{\hat {\mathcal{S}}_g\setminus \{j\}}\right) \frac{1}{n}\bx_j^T(\bI_n-\Pi_{\hat {\mathcal{S}}_g \setminus \{j\}})\bx_j.
\end{align*}
By Condition (A4), there exists a constant $\alpha>0$ such that
  \begin{align*}
\frac{1}{n}\bx_j^T(\bI_n-\Pi_{\hat {\mathcal{S}}_g \setminus \{j\}})\bx_j \geq \alpha.
\end{align*}
Therefore,
  \begin{align*}
\min_{j \in \hat {\mathcal{S}}_g \cap\mathcal{M}_{0,n}, ~g=1, \ldots, G_n}\Big|\frac{1}{n}\beta_j\bx_j^T(\bI_n-\Pi_{\hat{\mathcal{S}}_g \setminus \{j\}})\bx_j\Big|\geq \alpha\min_{j \in \mathcal{M}_{0, n}}|\beta_j|.
\end{align*}

{\bf Step 1.2} We next consider the second term in \eqref{eq:5} and show  that for $j =1, \dots, p_n$ and some $g$ such that $j \in \hat {\mathcal{S}_g}$,
 \begin{align*}
\Bigg|\sum_{k \in \mathcal{M}_{0,n} \setminus \hat{\mathcal{S}}_g} \frac{\beta_k}{n}\bx_j^T(\bI_n-\Pi_{\hat{\mathcal{S}}_g \setminus \{j\}})\bx_k\Bigg| \leq O(n^{-(b-a-d/2)}).
\end{align*}

Indeed,  by  the triangular inequality
\begin{align*}
\Bigg|\sum_{k \in \mathcal{M}_{0,n} \setminus \hat{\mathcal{S}}_g} \frac{\beta_k}{n}\bx_j^T(\bI_n-\Pi_{\hat{\mathcal{S}}_g \setminus \{j\}})\bx_k\Bigg| \leq \Bigg|\sum_{k \in \mathcal{M}_{0,n} \setminus \hat{\mathcal{S}}_g} \frac{\beta_k}{n} \bx_j^T\bx_k\Bigg|+ \Bigg|\sum_{k \in \mathcal{M}_{0,n} \setminus \hat{\mathcal{S}}_g} \frac{\beta_k}{n}\bx_j^T\Pi_{\hat{\mathcal{S}}_g \setminus \{j\}}\bx_k\Bigg|.
\end{align*}
By Condition (A1), $||\mathcal{M}_{0,n}||_0=O(n^a)$. By the construction of $\hat{\mathcal{S}}_g$,
\begin{align*}
|{n}^{-1}\bx_j^T\bx_k|\leq \delta_n
\end{align*}
for $j \in \hat{\mathcal{S}}_g$ and $k \in \mathcal{M}_{0,n} \setminus \hat{\mathcal{S}}_g$,  where $\delta_n=O(\sqrt{\log(p_n)/n})\leq O(n^{-b})$. By Condition (A5), $\max_{j \in  \mathcal{M}_{0,n}}|\beta_j|<M$. Therefore,
   \begin{align}
\Bigg|\sum_{k \in \mathcal{M}_{0,n} \setminus \hat{\mathcal{S}}_g} \frac{\beta_k}{n}\bx_j^T\bx_k\Bigg|\leq O(n^{-(b-a)}). \label{eq:6}
\end{align}
Moreover,
\begin{align*}
\Bigg|\sum_{k \in \mathcal{M}_{0,n} \setminus \hat{\mathcal{S}}_g} \frac{\beta_k}{n} \bx_j^T\Pi_{\hat{\mathcal{S}}_g \setminus \{j\}}\bx_k\Bigg|\leq \frac{Mn^a}{n}||\Pi_{\hat{\mathcal{S}}_g \setminus \{j\}}\bx_j||_2 \max_{k \in \mathcal{M}_{0,n} \setminus \hat{\mathcal{S}}_g}||\Pi_{\hat{\mathcal{S}}_g \setminus \{j\}}\bx_k||_2,
\end{align*}
where $||u||_2$ is the $\ell_2-$norm for $u \in \mathbb{R}^n$. By Condition (A4),
\begin{align*}
\lambda_{max}((\frac{1}{n}\bx_{\hat {\mathcal{S}}_g \setminus \{j\}}^T\bx_{\hat {\mathcal{S}}_g \setminus \{j\}})^{-1})\leq 1/\tau_{min}
\end{align*}
and hence
\begin{align*}
\max_{k \in \mathcal{M}_{0,n} \setminus \hat{\mathcal{S}}_g}\frac{1}{\sqrt{n}}||\Pi_{\hat{\mathcal{S}}_g \setminus \{j\}}\bx_k||_2 \leq O(n^{-(b-d/2)}).
\end{align*}
Moreover,
\begin{align*}
 \frac{1}{\sqrt{n}}||\Pi_{\hat{\mathcal{S}}_g \setminus \{j\}}\bx_j||_2 \leq \frac{1}{\sqrt{n}}||\bx_j||_2=1.
 \end{align*}
Thus, we have
\begin{align}
\Bigg|\sum_{k \in \mathcal{M}_{0,n} \setminus \hat{\mathcal{S}}_g} \frac{\beta_k}{n} \bx_j^T\Pi_{\hat{\mathcal{S}}_g \setminus \{j\}}\bx_k\Bigg| \leq O(n^{-(b-a-d/2)}). \label{eq:7}
\end{align}
Combining \eqref{eq:6} and \eqref{eq:7},
 \begin{align*}
\Bigg|\sum_{k \in \mathcal{M}_{0,n} \setminus \hat{\mathcal{S}}_g} \frac{\beta_k}{n}\bx_j^T(\bI_n-\Pi_{\hat{\mathcal{S}}_g \setminus \{j\}})\bx_k\Bigg| \leq O(n^{-(b-a-d/2)}).
\end{align*}

{\bf Step 1.3} We  move on to study the third term in \eqref{eq:5} and show that, under Condition (A6) and (A7), the third term is negligible compared to the first term. To proceed, we first reproduce a result from Lemma 14.9 of
\cite{Buhlmann2011} for the sake of readability.

 \textbf{Lemma 3 (Bernstein's inequality)}
{\it
Assume Condition (A6) and (A7). Let $t > 0$ be an arbitrary constant. Then
\[
P \left(\frac{1}{n}\sum_{i=1}^n \epsilon_i \geq K_{\epsilon}t+\sqrt{2t}\right)\leq\exp(-nt).
\]}

The following Lemma provides the ground for the proof of Step 1.3.

 \textbf{Lemma 4}
{\it
Assume Condition (A6) and (A7). For $t > 0$
\[
P \left(\max_{1\leq j \leq p_n} \frac{1}{n}\left|\bx_j^T\boldsymbol\epsilon\right| \geq K_0 \left(t+\frac{\log(2p_n)}{n}\right)+\sigma_0\sqrt{2\left(t+\frac{\log(2p_n)}{n}\right)}\right)\leq\exp(-nt),
\]}
where $K_0=KK_{\epsilon}$ and $\sigma_0=K\sigma$.

{\bf Proof of Lemma 4}

We have
 \begin{align*}
\frac{1}{n}\sum_{i=1}^n E|\epsilon_iX_{i, j}|^m \leq K^{m} \frac{1}{n} \sum_{i=1}^n E(|\epsilon_i|^m)
\leq\frac{m!}{2} (KK_{\epsilon})^{m-2}(K\sigma)^2=\frac{m!}{2} (K_0)^{m-2}(\sigma_0)^2, ~m=2,3, \ldots.
\end{align*}
Therefore, $\epsilon_iX_{i, j}$ follows a sub-exponential distribution as well.
Lemma 3 (Bernstein's inequality) implies that
  \begin{align*}
&{}P \left(\frac{1}{n}\left|\bx_j^T\boldsymbol\epsilon\right| \geq K_0 \left(t+\frac{\log(2p_n)}{n}\right)+\sigma_0\sqrt{2\left(t+\frac{\log(2p_n)}{n}\right)} \right)
\leq  2\exp\left(-n\left(t+\frac{\log(2p_n)}{n}\right)\right).
\end{align*}
Therefore,
  \begin{align*}
&{}P \left(\max_{1\leq j \leq p_n}\frac{1}{n}\left|\bx_j^T\boldsymbol\epsilon\right| \geq K_0 \left(t+\frac{\log(2p_n)}{n}\right)+\sigma_0\sqrt{2\left(t+\frac{\log(2p_n)}{n}\right)} \right)\\
\leq &{} \sum_{j=1}^{p_n} P \left(\frac{1}{n}\left|\bx_j^T\boldsymbol\epsilon\right| \geq K_0 \left(t+\frac{\log(2p_n)}{n}\right)+\sigma_0\sqrt{2\left(t+\frac{\log(2p_n)}{n}\right)} \right) \\
\leq &{} 2p_n\exp\left(-n\left(t+\frac{\log(2p_n)}{n}\right)\right) = \exp(-nt).
\end{align*}

{\bf Proof of Step 1.3}

We  move on to study the third term in \eqref{eq:5} and show that the third term is negligible compared to the first term. We have
 \begin{align*}
K_0 \left(n^{-2b}+\frac{\log(2p_n)}{n}\right)\leq K_0 \left(\frac{n^{1-2b}+\log(2p_n)}{n}\right) \leq 3K_0n^{-2b},
\end{align*}
 \begin{align*}
\sigma_0\sqrt{2 \left(n^{-2b}+\frac{\log(2p_n)}{n}\right)} \leq \sqrt{6}\sigma_0n^{-b}.
\end{align*}
Therefore,
  \begin{align*}
  &{}P \left(\max_{1\leq j \leq p_n} \frac{1}{n}\left|\bx_j^T\boldsymbol\epsilon\right| \geq
  3K_0n^{-2b}+\sqrt{6}\sigma_0n^{-b}\right)\\
\leq &{}P \left(\max_{1\leq j \leq p_n} \frac{1}{n}\left|\bx_j^T\boldsymbol\epsilon\right| \geq K_0 \left(n^{-2b}+\frac{\log(2p_n)}{n}\right)+\sigma_0\sqrt{2\left(n^{-2b}+\frac{\log(2p_n)}{n}\right)}\right)
\leq  \exp(-n^{1-2b}),
\end{align*}
where the last inequality holds by Lemma 4.
Similarly
\begin{align*}
P\left(\max_{1\leq j\leq p_n}\frac{1}{n}|\bx_j^T(\bI_n-\Pi_{\hat{\mathcal{S}}_g \setminus \{j\}})\boldsymbol\epsilon| \leq 3K_0n^{-2b}+\sqrt{6}\sigma_0n^{-b} \right)
 \geq 1-\exp\left(-n^{1-2b}\right).
\end{align*}

{\bf Step 1.4} We are now in a position to show the first statement in Theorem 1. Indeed, we have so far shown that for $j \in \mathcal{M}_{0,n}$, the last two terms in
\eqref{eq:5}  are negligible compared to the first term. Similarly, the two terms in the denominator  of $\hat\rho(Y_i,X_{i,j}|\bX_{i,\hat {\mathcal{S}}_g \setminus \{j\}})$ can be shown to be bounded from above by Conditions (A4) and (A5), respectively. Combining these results and applying the Bonferroni inequality, we have
 \begin{align}
P\left(\min_{j \in \hat {\mathcal{S}}_g \cap\mathcal{M}_{0,n}, ~g=1, \ldots, G_n} \nu_n^{-1}|\hat\rho(Y_i,X_{i,j}|\bX_{\hat {\mathcal{S}}_g \setminus \{j\}})| \geq O(n^{\kappa}\min_{j \in  \mathcal{M}_{0,n}}|\beta_j|)\right) \geq 1-O\left(\exp\left(-n^{1-2b}\right)\right). \label{eq:8}
\end{align}
That is, the CIS procedure satisfies the sure screening property
  \begin{align*}
P(\mathcal{M}_{0,n} \subseteq \widehat{\mathcal{M}}_{CIS}(\nu_{n}, \delta_n)) \geq 1-O\left(\exp\left(-n^{1-2b}\right)\right).
\end{align*}

{\bf Step 2: $j \in \{1, \dots, p_n\} \setminus \mathcal{M}_{0,n}$}

We then move on to prove the second statement in Theorem 1 by considering the scenario  when $j \in \{1, \dots, p_n\} \setminus \mathcal{M}_{0,n}$. Since $\beta_j=0$ for $j \in \{1, \dots, p_n\} \setminus \mathcal{M}_{0,n}$, the first term in
\eqref{eq:5}  vanishes. Also, we showed that the absolute values of the second and the third terms can be bounded from above.  Coupled with the fact that the denominator of $\hat\rho(Y_i,X_{i,j}|\bX_{i,\hat {\mathcal{S}}_g \setminus \{j\}})$ is bounded from below, an application of the   Bonferroni inequality yields
 \begin{align}
&{} P\Bigg(\max_{j \in \hat {\mathcal{S}}_g \setminus\mathcal{M}_{0,n}, g=1, \dots, G_n}\nu_n^{-1}|\hat\rho(Y_i,X_{i,j}|\bX_{i,\hat {\mathcal{S}}_g \setminus \{j\}})|\leq O(n^{-(b-a-d/2-\kappa)})\Bigg)
&{} \geq 1-O\left(\exp\left(-n^{1-2b}\right)\right). \label{eq:9}
\end{align}

Finally, the first and the second statements in Theorem 1  immediately imply the screening consistency property:
  \begin{align*}
P(\widehat{\mathcal{M}}_{CIS}(\nu_{n}, \delta_n)=\mathcal{M}_{0,n} ) \geq 1-O\left(\exp\left(-n^{1-2b}\right)\right).
\end{align*}
\bigskip
\begin{center}
{\large\bf SUPPLEMENTARY MATERIAL}
\end{center}

\noindent
Example R codes are contained in the zip file {\it CIS.zip} available online. An R package will soon be uploaded to the CRAN repository. Additional technical details referenced in Sections 2-4, technical proofs for Lemma 1-2 and Web Figure A1 can be found in the Web\_Supp.pdf file. The complete data set can be downloaded from The Cancer Genome Atlas ({\it https://cancergenome.nih.gov/}).

\bigskip
\begin{center}
{\large\bf ACKNOWLEDGEMENT}
\end{center}

\noindent
The work was partially  supported by grants from the NSA  (H98230-15-1-0260: Hong) and the National Natural Science Foundation of China (No.11528102: Lin and Li).


\bibliographystyle{chicago}

\bibliography{Bibliography-CIS}

\newpage

\begin{figure}[b]
 \caption{\label{fig:solution} (a) Graphical representation of 2,339 SNPs shown in He et al. (2015).  The correlation plot clearly shows that SNPs form 15 distinct pathways. Specifically, SNPs are placed in the same pathway when their absolute sample correlation is moderate or strong ($ \ge 0.2$). (b)
15 pathways presented in (a) can be presented by a block-diagonal covariance matrix.}
\centering
\subfloat[Correlation Plot]{
\includegraphics[width=2.5in, height=2in, angle=180]{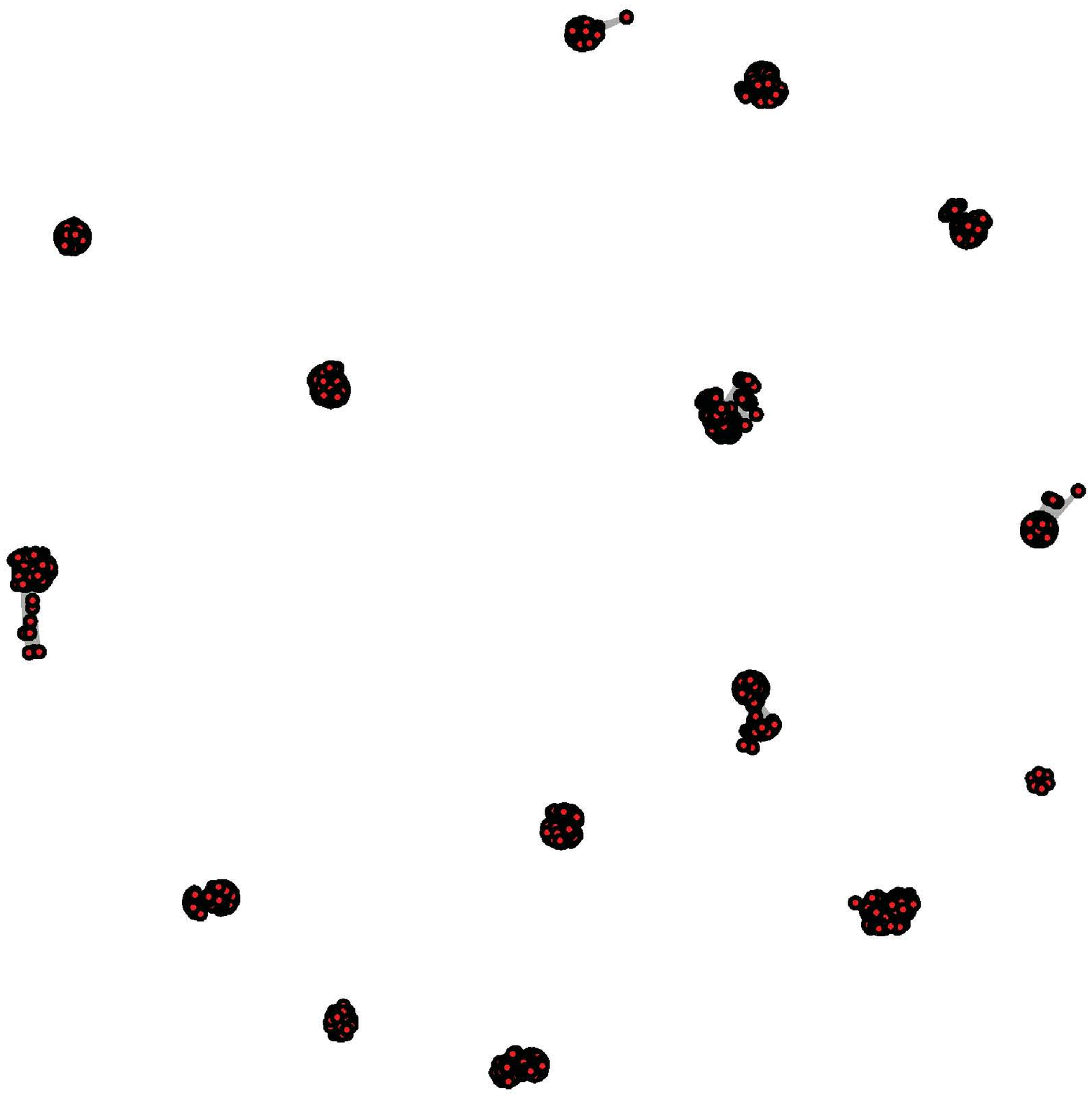}}
     \centering
\vspace{1pt}
\subfloat[Correlated Blocks]{
\includegraphics[width=2in, height=2.5in, angle=270]{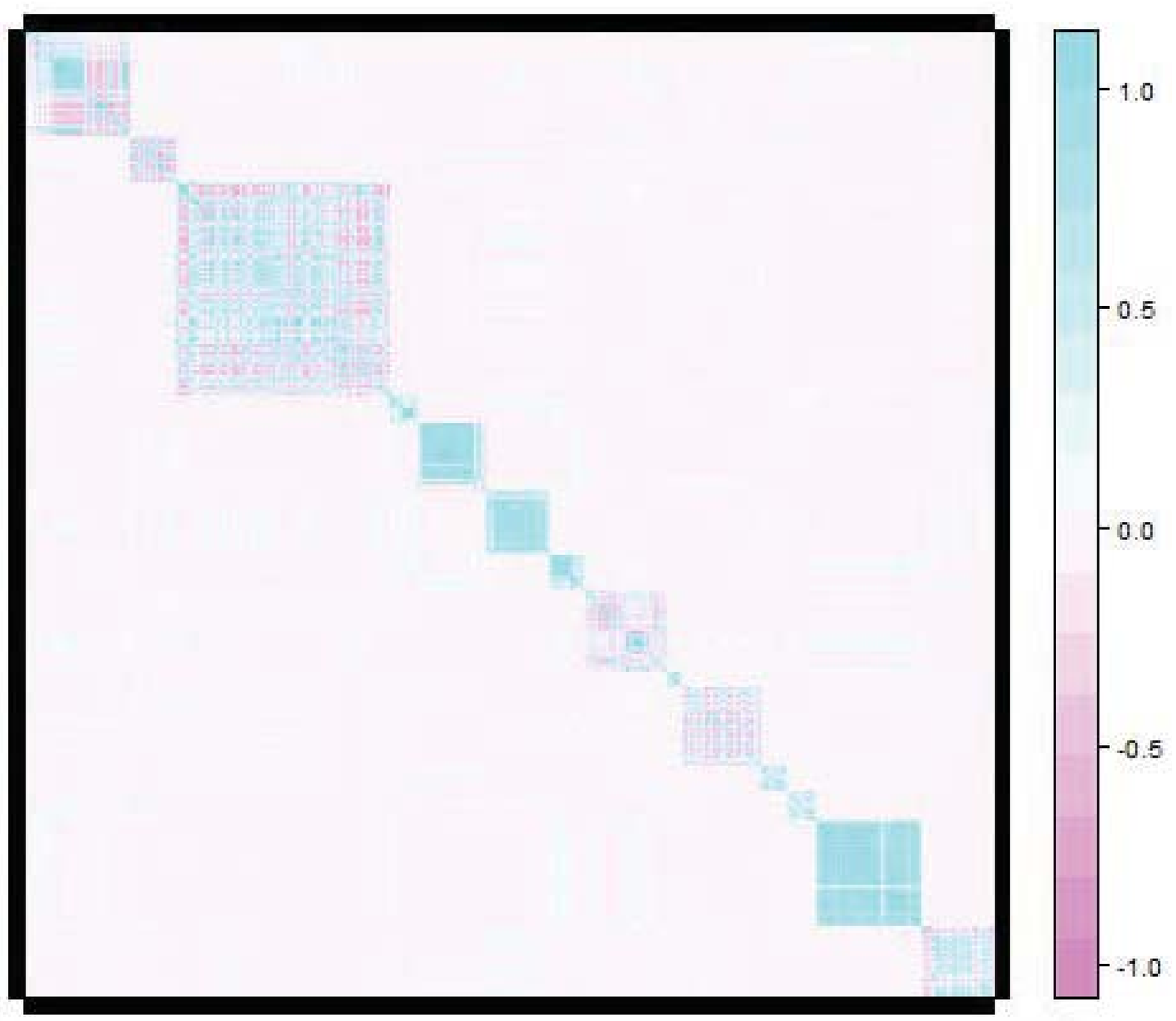}}
\end{figure}

\begin{figure}[b]
 \caption{(a) The correlation block containing PLCE1, EHD2, LINC00665 and ZNF295-AS1. (b)
Correlation plot}
\centering
\subfloat[Correlated Block]{
\includegraphics[width=2.5in, height=2.5in]{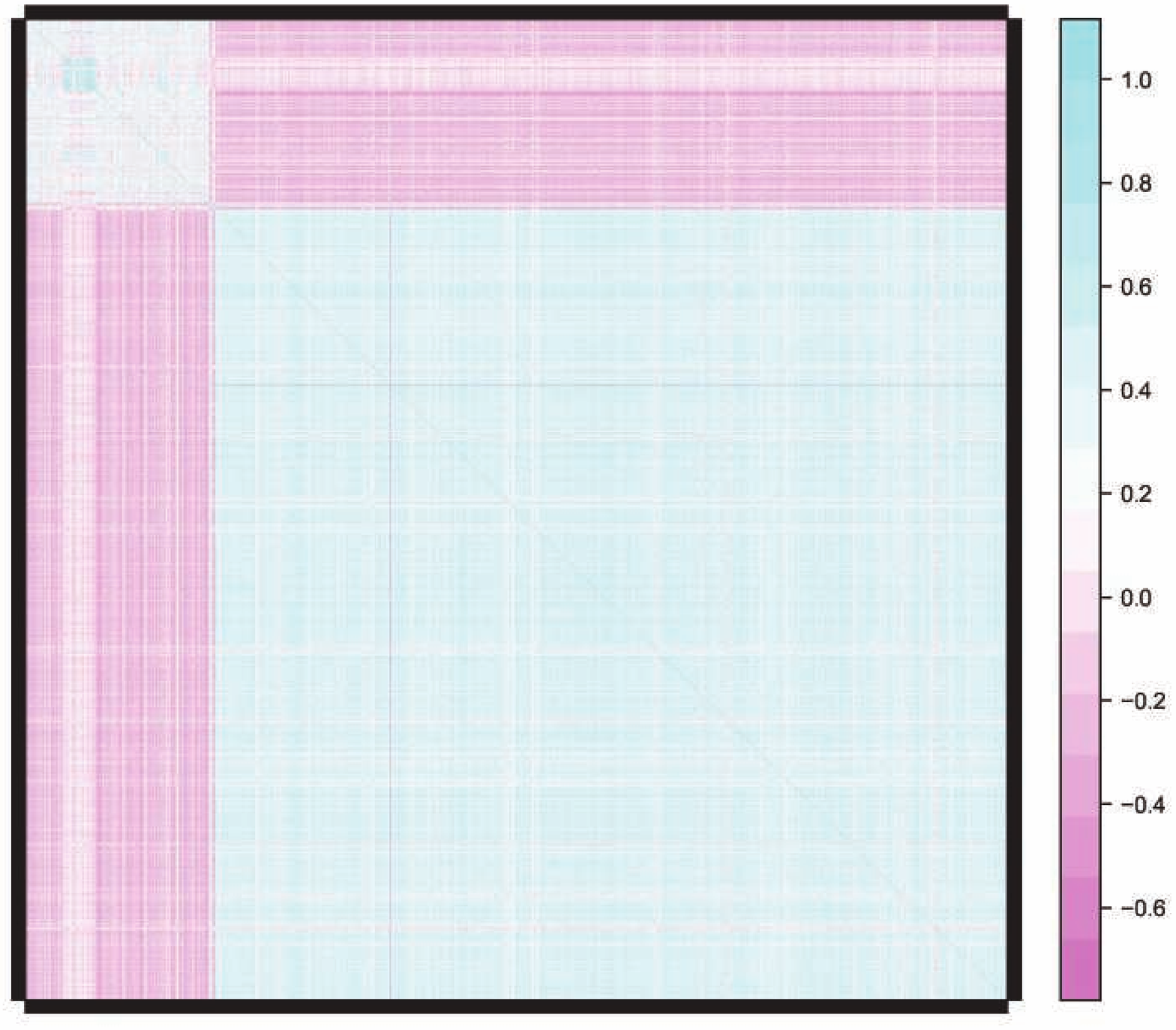}}
     \centering
\vspace{1pt}
\subfloat[Correlation Plot]{
\includegraphics[width=2.5in, height=2.5in]{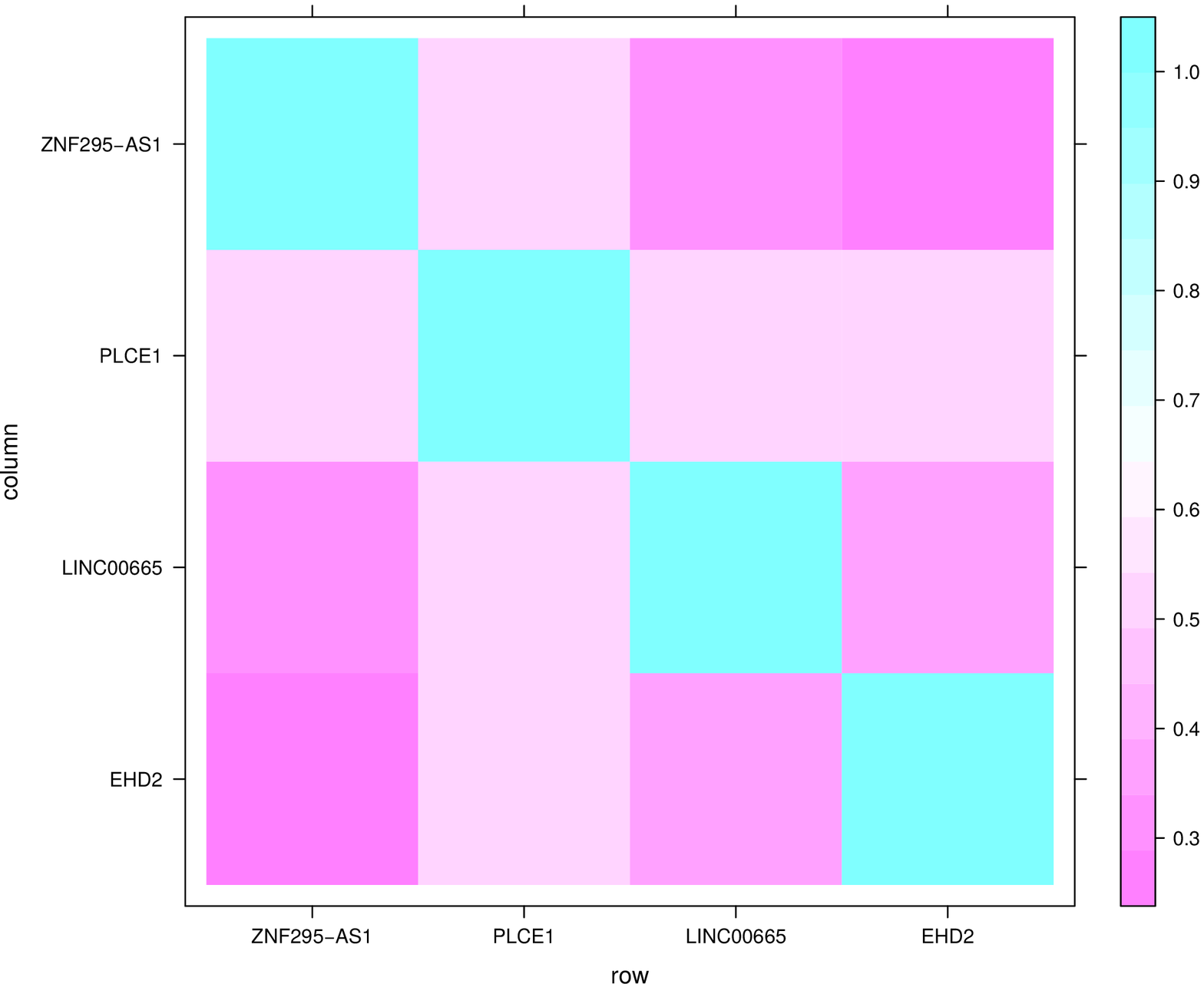}}
\end{figure}

\begin{table}[b] 
\small
\caption{Average number of selected variables required to include the true  model (standard deviation in parentheses) for Models A-C.}
\begin{center}
\begin{tabular}{cc rrrr}
\hline
 Models &   Correlation & SIS & HOLP & Tilting & CIS  \\
 \hline
  A   & 0.9 &  7103.2 (1937.4)  & 3458.4 (2932.2) & 1041.1 (648.7) & 73.7 (141.7) \\
           & 0.8  &      2835.0 (2334.0) & 660.7 (877.2)  & 168.3 (131.2)  & 11.4 (5.1) \\
         &   0.7 &      505.9 (594.9) & 102.9 (253.2) & 31.0 (8.7)   & 10.0 (0.0) \\
       &  0.6 &      60.4 (75.3) & 21.9 (20.9) & 21 (2.9)  & 10.0 (0.0) \\
          & 0.5  &     18.1 (7.7) & 13.2 (2.2) & 19.5 (21.7)  & 10.0 (0.0) \\
  \hline
  B   & 0.9  &  21.2 (5.4) & 12.2 (1.9)  & 154.2 (515.1) & 68.6 (80.4) \\
           & 0.8  &      13.2 (2.5) & 10.7 (1.1) & 13.9 (4.4)  & 10.9 (2.2) \\
         &   0.7 &     10.7 (1.1) & 10.2 (0.5) & 10.7  (1.1) & 10.0 (0.0) \\
     &   0.6 &     10.2 (0.5) & 10.0 (0.0) & 10.2 (0.5)   & 10.0 (0.0)\\
          & 0.5  &      10.0 (0.0) & 10.0 (0.0)& 10.0  (0.0) & 10.0 (0.0)\\
  \hline
   C   & 0.9 &  3729.2 (944.3) & 220.7 (555.1) & 910.4 (1215.3)  & 270.3 (453.4)  \\
           & 0.8 &     1570.4 (1343.2) & 47.2 (946.1) & 147.8 (86.2)   & 14.8 (8.6) \\
         &   0.7 &     350.6 (523.2) & 27.0 (11.6) & 47.4 (16.5)  & 10.8 (0.8) \\
     &   0.6 &     58.1 (54.5) & 17.8 (3.3) & 26.5 (5.7)  & 10.2 (0.4) \\
          & 0.5 &      21.9 (5.0) & 11.5 (1.7) & 18.3  (3.9)  & 10.0 (0.1) \\
  \hline
\end{tabular}
\end{center}
\end{table}

\begin{table}[b] 
\small
\caption{Numbers of false positive (FP) and numbers of false negative (FN) for Model D.}
\begin{center}
\begin{tabular}{cclccccccc}
\hline
   p & $|\beta|$ & Measures  & Lasso &  Adaptive Lasso  & ISIS & HOLP & Tilting   &  ICIS \\
  \hline
    10,000     & 0.5 & FP & 45.47 & 0.01 & 0.02 & 0.13 &NA & 0.20    \\
          & & FN & 3.47 & 4.00& 4.00  &4.00  &NA  &0.33 \\
            & & FP+FN & 48.94 & 4.01 & 4.02 &  4.13  &NA  &0.53 \\
                 \hline
            & 0.75 & FP & 145.57 & 0.10 & 26.88  & 0.19 &NA  &0.22  \\
                 & & FN & 1.31 & 2.31 & 2.85 & 4.00  & NA  &0.00  \\
             & & FP+FN & 146.88 & 2.43 & 29.63  & 4.19  &NA &0.22  \\
                             \hline
  & 1 & FP & 220.87 & 0.06 & 23.62  & 0.21 &NA  &0.08  \\
                 & & FN & 0.00 & 0.00 & 0.14 & 4.00  & NA  &0.00  \\
             & & FP+FN & 220.87 & 0.06 & 23.76  & 4.21  &NA &0.08  \\
                             \hline
    1,000       & 0.5 & FP & 85.35 & 7.64 & 0.80 & 0.27 &0.21 & 0.73    \\
          & & FN & 0.28 & 0.52& 3.80  &3.09  &3.08   &0.06 \\
            & & FP+FN & 85.83 & 8.16 & 4.60 &  3.36  &3.29  &0.79 \\
                 \hline
            & 0.75 & FP & 90.85 & 3.83 & 10.15  & 0.75 &0.95  &0.80  \\
                 & & FN & 0.00 & 0.00 & 0.12  & 0.32 & 0.90  &0.00  \\
             & & FP+FN & 90.85 & 3.83 & 10.27  & 1.07  &1.85 &0.80  \\
            \hline
              & 1 & FP & 92.49 & 2.28 & 2.15  & 0.13 &0.78  &0.22  \\
                 & & FN & 0.00 & 0.00 & 0.00  & 0.00 & 0.00  &0.00  \\
             & & FP+FN & 92.49 & 2.28 & 2.15  & 0.13  &0.78 &0.22  \\
  \hline
\end{tabular}
\end{center}
\end{table}

\begin{table}[b] 
\caption{Numbers of false positive (FP) and false negative (FN) for Model E.}
\begin{center}
\begin{tabular}{cccccc}
\hline
    Measures & Lasso &  Adaptive Lasso & ISIS & GS & ICIS\\
 \hline
   FP     & 959.84 &0.16  & 137.90 & 48.78 & 7.51 \\
         FN    & 2.99  &21.66  & 10.90  & 26.04  & 5.55\\
          FP+FN   & 962.83  &21.82  & 148.80 & 74.82 & 13.06\\
  \hline
\end{tabular}
\end{center}
\end{table}

\end{document}